\documentclass{article}
\usepackage{spconf,amsmath,graphicx}
\usepackage{graphicx}
\usepackage{amssymb,amsmath,bm}
\usepackage{textcomp}
\usepackage{multirow}
\usepackage{threeparttable}
\usepackage{comment}
\usepackage{psfrag}
\usepackage{color}
\usepackage{epsfig}
\usepackage{epstopdf}
\usepackage{pstool}

\title{Future Word Contexts in Neural Network Language Models}

 \name{ {\em X. Chen$^1$, X. Liu$^2$, A. Ragni$^1$, Y. Wang$^1$, M.J.F. Gales$^1$}
\thanks{This research was funded under the ALTA Institute, University of Cambridge.
Thanks to Cambridge English, University of Cambridge, for supporting this research.
Xunying Liu is funded by MSRA grant no. 6904412 and CUHK grant no. 4055065.
}}
\address{
    { University of Cambridge Engineering Department $^1$,} 
      Chinese University of Hong Kong $^2$ \\
 { \small\tt \{xc257,ar527,yw396,mjfg\}@eng.cam.ac.uk, xyliu@se.cuhk.edu.hk}
}
\begin{document}
\ninept
\maketitle
\begin{abstract}
Recently, bidirectional recurrent network language models (bi-RNNLMs) have been shown to outperform standard,
unidirectional, recurrent neural network language models (uni-RNNLMs) on a range of speech recognition tasks. This indicates 
that future word context information beyond the word history can be useful. However, bi-RNNLMs pose a number
of challenges as they make use of the complete previous and future word context information. This impacts both training
efficiency and their use within a lattice rescoring framework. In this paper these issues are addressed by proposing
a novel neural network structure, succeeding word RNNLMs (su-RNNLMs). Instead of using a recurrent
unit to capture the complete future word contexts, a 
feedforward unit is used to model a finite  number of succeeding, future, words. This model can be trained much more 
efficiently than bi-RNNLMs and can also be used for lattice rescoring. 
Experimental results on a meeting transcription
task (AMI)  show the proposed model consistently outperformed uni-RNNLMs and 
yield only a slight degradation compared to bi-RNNLMs in N-best rescoring.
Additionally,
performance improvements can be obtained using lattice rescoring and subsequent confusion network decoding.
\end{abstract}
\begin{keywords}
Bidirectional recurrent neural network, language model, succeeding words, speech recognition
\end{keywords}
\section{Introduction}
\label{sec:intro}

Language models (LMs) are crucial components in many applications, such as
speech recognition and machine translation. The aim of language models
is to compute the probability of any given sentence $\mathcal{W}=(w_1, w_2, ..., w_L)$,
which can be calculated as
\begin{equation}
P(\mathcal{W}) = P(w_1, w_2, ..., w_L) = \prod_{t=1}^{L} P(w_t | w_1^{t-1})
\end{equation}
The task of LMs is to calculate the probability of word $w_t$ given its previous history 
$w_1^{t-1}=w_1, w_2, ...,w_{t-1}$.
$n$-gram LMs \cite{chen1999LMsmoothing}  and neural network based language mdoels (NNLMs) \cite{Benjio:2003,RNNLM}
are two  widely used language models.
In $n$-gram LMs, the most recent $n-1$ words are used 
as an approximation of the complete history, thus
\begin{equation}
P(w_t | w_1^{t-1}) \approx P(w_t | w_{t-n+1}^{t-1})
\end{equation}
This $n$-gram assumption can also be used to construct a $n$-gram feedforward NNLMs \cite{Benjio:2003}.
In contrast, recurrent neural network LMs (RNNLMs) model the complete history via a recurrent connection.
% They have been reported to achieve state-of-the-art performance in a range of tasks \cite{RNNLM}.
% In this paper, this form of language model which only uses history context information is referred
% as unidirectional language models (uni-LMs)

Most of previous work on language models has focused on utilising history information,
the future word context information has not been extensively investigated.
There have been several attempts to incorporate future  context information into recurrent neural network language models.
Individual forward and backward RNNLMs can be built, and these two LMs combined
with a log-linear interpolation \cite{xiong2016achieving}. In \cite{shi2013interspeech}, succeeding words were incorporated into 
RNNLM within a Maximum Entropy framework. \cite{arisoy2015bidirectional} 
investigated the use of bidirectional RNNLMs (bi-RNNLMs) for speech recognition.
For a broadcast news task, sigmoid based RNNLMs gave small gains, while no performance improvement was obtained when using 
long short-term memory (LSTM) based RNNLMs.
More recently,  bi-RNNLMs can produce consistent, and significant, performance improvements
over unidirectional RNNLMs (uni-RNNLMs) on a range of speech recognition tasks \cite{xie2017interspeech}. 

Though they can yield performance gain, bi-RNNLMs pose several challenges for
both model training and inference as they require the complete previous and
future word context information to be taken into account. It is difficult to parallelise training efficiently.
Lattice rescoring is also complicated for these LMs as future context needs to be incorporated.
This means that the form of approximation used for uni-RNNLMs \cite{liu2016taslp} is not suitable to apply.
Hence, N-best rescoring is normally used \cite{shi2013interspeech, 
arisoy2015bidirectional, xie2017interspeech}. 
However, the ability to manipulate lattices
is very important in many speech applications. Lattices can be used for a wide range of downstream applications,
such as confidence score estimation \cite{wessel2001confidence}, 
keyword search \cite{chen2017icassp} and confusion network decoding \cite{CN}.
In order to address these issues, a novel model structure, succeeding word RNNLMs (su-RNNLMs), is proposed in this paper.
Instead of using a recurrent unit to capture the complete 
future word context as in bi-RNNLMs, a feedforward unit 
is used to model a small, fixed-length number of succeeding words.
This allows existing efficient training \cite{chen2016taslp} and lattice rescoring
\cite{liu2016taslp} algorithms developed for uni-RNNLMs
to be extended to the proposed su-RNNLMs.
Using these extended algorithms, compact lattices can be generated with su-RNNLMs
supporting lattice based downstream processing.
% and consistent performance improvements are
% obtained by subsequent confusion network decoding \cite{CN}.

The rest of this paper is organized as follows. Section \ref{sec:rnnlms} gives a brief review
of RNNLMs, including both unidirectional and bidirectional RNNLMs.
The proposed model with succeeding words (su-RNNLMs) is introduced in Section \ref{sec:rnnlmsuccwords},
followed by a description of the lattice rescoring algorithm in Section \ref{sec:rnnlmsuccwordslatrescore}.
Section \ref{sec:intpltlms} discusses the interpolation of language models.
The experimental results are presented in Section \ref{sec:exps} 
and conclusions are drawn in Section \ref{sec:conclusion}.

\section{Uni- and Bi-directional RNNLMs}
\label{sec:rnnlms}
\subsection{Unidirectional RNNLMs}
\begin{figure}
   \includegraphics[width=8.0cm]{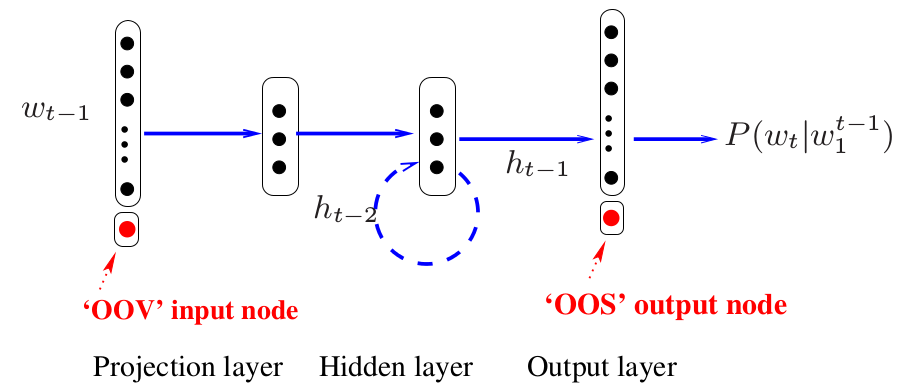}
   \caption{{\it An example unidirectional RNNLM.}}
\label{fig:unirnnlm}
 \end{figure}

In contrast to feedforward NNLMs, where only modeling the previous $n-1$ words,
 recurrent NNLMs \cite{NNLM} represent
the full non-truncated history $w_1^{t-1}=w_1, w_2, ..., w_{t-1}$ for word
$w_t$ using the 1-of-K encoding of the previous word $w_{t-1}$ and a continuous
vector $h_{t-2}$ as a compact representation of the remaining context $w_1^{t-2}$.
Figure \ref{fig:unirnnlm} shows an example of this unidirectional RNNLM (uni-RNNLM).
The most recent word $w_{t-1}$ is used as input and
 projected into a low-dimensional, continuous, space via a linear projection layer.
A recurrent hidden layer is used after this projection layer.
The form of the recurrent layer can be based on a standard sigmoid based recurrent unit, with sigmoid
activations \cite{RNNLM}, or more complicated forms such as gated recurrent unit 
(GRU) \cite{chung2014gru} and long short-term memory (LSTM) units \cite{hochreiter1997long}.
A continuous vector $h_{t-1}$ representing the complete history information $w_1^{t-1}$ can be 
obtained using $h_{t-2}$ and previous word $w_{t-1}$. This vector is used 
as input of recurrent layer for the estimation of next word.
An output layer with softmax function is used to calculate the probability $P(w_{t} | w_{1}^{t-1})$.
An additional node is often added at the output layer to model the probability mass of
out-of-shortlist (OOS) words to speed up softmax computation by limiting vocabulary size
~\cite{park2010improved}. Similarly, 
an out-of-vocabulary (OOV) node can be added in the input layer to model OOV words. 
The probability of word sequence $\mathcal{W}=w_1^L$ is calculated as,
\begin{equation}
 P_{u} (w_1^L) = \prod_{t=1}^{L} P(w_t | w_1^{t-1})
 \label{eqn:uni}
\end{equation}

Perplexity (PPL) is a metric used widely to evaluate the quality of language models.
According to the definition in \cite{jelinek2009cl}, the perplexity can
be computed based on sentence probability with,
\begin{eqnarray}
\label{eqn:unippl}
{\tt PPL} &=& \exp\big( - \frac{1}{N} \sum_{j=1}^{J} \log P_u(\mathcal{W}_{j}) \big) \nonumber \\
       &=& \exp\big( - \frac{1}{N} \sum_{j=1}^{J} \log P_u(w_1^{L_j}) \big) \nonumber \\
       &=&  \exp \big( - \frac{1}{N} \sum_{j=1}^{J} \sum_{t=1}^{L_{j}} \log P(w_t | w_1^{t-1}) \big)
\end{eqnarray}
Where $N$ is the total number of words and $J$ is the number of sentence in the 
evaluation corpus. $L_j$ is the number of word in $j$th sentence. From the above equation, 
the PPL is calculated based on the average log probability of each word, which for unidirectional
LMs, yields the average sentence log probability.

Uni-RNNLMs can be trained efficiently on Graphics Processing Units (GPUs) 
by using spliced sentence bunch (i.e. minibatch) mode \cite{chen2016taslp}. 
% Cross entropy is normally used as the training criterion.
Multiple sentences can be concatenated together to form a longer sequence and sets of these long sequences
can then be aligned in parallel from left to right. This data structure is more efficient for minibatch 
based training as they have comparable sequence length \cite{chen2016taslp}.
When using these forms of language models for tasks like speech recognition, 
N-best rescoring is the most straightforward way to apply uni-RNNLMs.
Lattice rescoring is also possible by introducing approximations \cite{liu2016taslp}
to control merging and expansion of different paths in lattice. This will be 
described in more detail in Section \ref{sec:rnnlmsuccwordslatrescore}.

\subsection{Bidirectional RNNLMs}
\label{sec:birnnlms}
\begin{figure}
   \includegraphics[width=8.0cm]{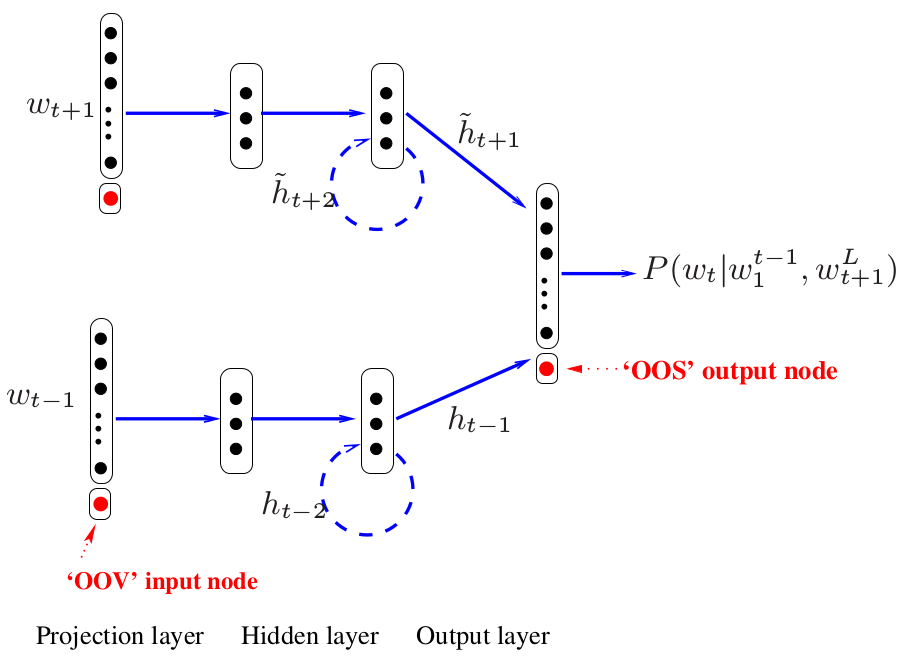}
   \caption{{\it An example bidirectional RNNLM.}}
\label{fig:birnnlm}
 \end{figure}

Figure \ref{fig:birnnlm} illustrates an example of bidirectional RNNLMs (bi-RNNLMs).
Unlike uni-RNNLMs, both the history word context $w_{1}^{t-1}$ and future word context $w_{t+1}^L$ are used
to estimate the probability of current word $P(w_t |w_{1}^{t-1}, w_{t+1}^L)$. 
Two recurrent units are used to capture the previous and future
 information respectively.
In the same fashion as uni-RNNLMs, $h_{t-1}$ is a compact continuous vector of
the history information $w_{1}^{t-1}$. While $\tilde{h}_{t+1}$ is another 
continuous vector to encode the future information $w_{t+1}^{L}$.
This future context vector is computed from 
the next word $w_{t+1}$ and the previous future context vector $\tilde{h}_{t+2}$ containing 
information of $w_{t+2}^{L}$.
 The concatenation of $h_{t-1}$ and $\tilde{h}_{t+1}$
 is then fed into the output layer, with softmax function, to calculate the output probability.
In order to reduce the number of parameter, the projection layer for the previous and future words are often shared.

The probability of word sequence $\mathcal{W}=w_1^L$ can be computed using bi-RNNLMs
as,
\begin{equation}
 P_{b} (w_1^L) = \frac{1}{Z_{b}} \hat{P}_{b} (\mathcal{W})  = \frac{1}{Z_{b}} \prod_{t=1}^{L} P(w_t | w_1^{t-1}, w_{t+1}^L) 
  \label{eqn:bi}
\end{equation}
$\hat{P}_{b} (\mathcal{W}) $ is the unnormalized sentence probability computed from the
individual word probabilities of the bi-RNNLM.
$Z_{b}$ is a sentence-level normalization term to ensure the 
sentence probability is appropriately normalized. This is defined as,
\begin{equation}
 Z_{b} = \sum_{\mathcal{W} \in \Theta} \hat{P}_{b} (\mathcal{W})  
\label{eqn:birnnlmnorm}
\end{equation}
where
$\Theta$ is the set of all possible sentences. Unfortunately, this normalization
term is impractical to calculate for most tasks.

In a similar form to Equation \ref{eqn:unippl}, the PPL of bi-RNNLMs can be calculated based on sentence probability as,
\begin{eqnarray}
\label{eqn:bippl}
{\tt PPL} &=& \exp\big( - \frac{1}{N} \sum_{j=1}^{J} \log P_b(w_1^{L_j}) \big) \nonumber \\
&=& \exp\big( - \frac{1}{N} \sum_{j=1}^{J} \log \frac{1}{Z_b} \hat{P}_b(w_1^{L_j}) \big)  \\
&=& \exp\big(\frac{J}{N}\log(Z_b)  - \frac{1}{N} \sum_{j=1}^{J} \sum_{t=1}^{L_{j}} \log P(w_t | w_1^{t-1}, w_{t+1}^{L_{j}}) \big) \nonumber
\end{eqnarray}
However,  $Z_b$ is often infeasible to obtain. 
As a result, it is not possible to compute a valid perplexity from bi-RNNLMs. 
Nevertheless, the average log probability of each word can be used
to get a ``pseudo'' perplexity (PPL).
\begin{equation}
\label{eqn:bipplv1}
{\tt PPL_{pseudo} } =\exp\big(  - \frac{1}{N} \sum_{j=1}^{J} \sum_{t=1}^{L_{j}} \log P(w_t | w_1^{t-1}, w_{t+1}^{L_{j}}) \big)
\end{equation}
This is the second term of the valid PPL of bi-RNNLMs shown in Equation \ref{eqn:bippl}.
It is a ``pseudo'' PPL because the normalized sentence probability ${P}_{b} (\mathcal{W})$ 
is impossible to obtain and the unnormalized sentence probability $\hat{P}_{b} (\mathcal{W})$ is used instead.
Hence, the ``pseudo'' PPL of bi-RNNLMs is not comparable with the valid PPL of uni-RNNLMs.
However, the value of ``pseudo'' PPL provides information on the average word probability
from bi-RNNLMs since it is obtained using the word probability.

In order to achieve good performance for speech recognition,
 \cite{xie2017interspeech} proposed an additional smoothing of the bi-RNNLM probability
 at test time. The probability of bi-RNNLMs is smoothed as,
\begin{equation}
 P(w_i | w_1^{t-1}, w_{t+1}^{L}) = \frac{ \exp(\alpha y_i)}{\sum_{j}^{\mathcal{V}}  \exp(\alpha y_j)}
 \label{eqn:smooth}
\end{equation}
where  $y_i$ is the activation before softmax function for node $i$ in the output layer.
$\alpha$ is an empirical smoothing factor, which is chosen as 0.7 in this paper.

The use of both preceding and following context information in bi-RNNLMs
presents challenges to both model training and inference. First, 
N-best rescoring is normally used for speech recognition \cite{xie2017interspeech}.
Lattice rescoring is impractical for bi-RNNLMs as
the computation of word probabilities requires information from the complete sentence.

Another drawback of bi-RNNLMs is the difficulty in training.
The complete previous and future context information is required
 to predict the probability of each word. 
It is expensive to directly training bi-RNNLMs sentence by sentence, and
difficult to parallelise the training for efficiency.
In \cite{arisoy2015bidirectional}, all sentences in the training corpus were
concatenated together to form a single sequence to facilitate minibatch based training. This sequence was then
``chopped'' into sub-sequences with the average sentence
length. Bi-RNNLMs were then trained on GPU by
processing multiple sequences at the same time. This allows bi-RNNLMs to be
efficiently trained. However, issues can arise from the random
cutting of sentences, history and future context vectors may be reset in the middle
of a sentence. In \cite{xie2017interspeech}, the bi-RNNLMs were trained in a more
consistent fashion.  Multiple sentences were aligned from left to right
to form minibatches during bi-RNNLM training.  In order to handle
issues caused by variable sentence length, {\tt NULL} tokens were appended
to the ends of sentences to ensure that the aligned sentences had the
same length. These {\tt NULL} tokens were not used for parameter update.
In this paper,  this approach is adopted to train bi-RNNLMs as it gave better performance.
% Hence, it will be very slow to train bi-RNNLMs in an ``accurate'' way.

\section{RNNLMs with succeeding words}
\label{sec:rnnlmsuccwords}
\begin{figure}
   \includegraphics[width=8.0cm]{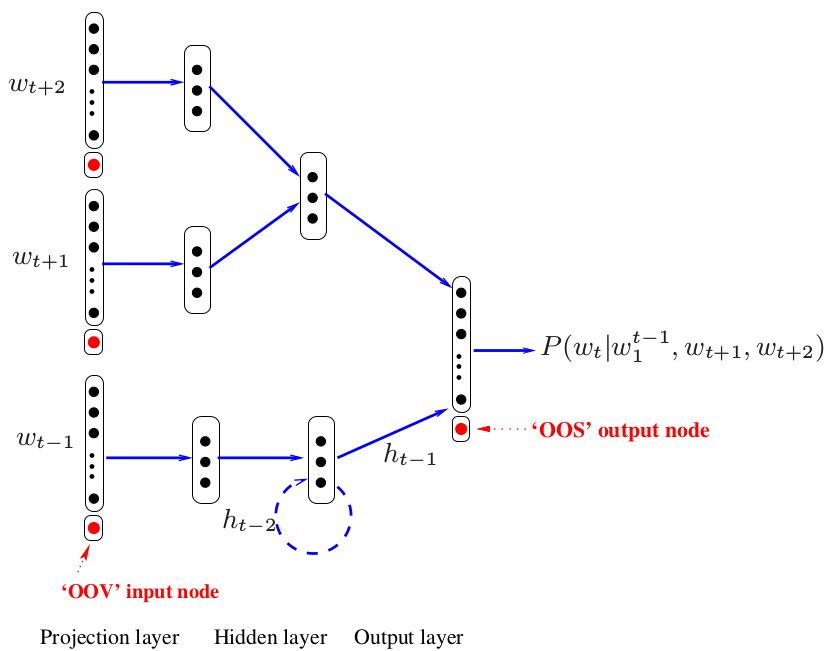}
   \caption{{\it An example su-RNNLM with 2 succeeding words.}}
\label{fig:rnnlmsw}
 \end{figure}

As discussed above, bi-RNNLMs are slow to train and difficult to use in lattice rescoring.
In order to address these issues, a novel structure, the su-RNNLM, is proposed in this paper 
to incorporate future context information.
The model structure is illustrated in Figure \ref{fig:rnnlmsw}.
In the same fashion as bi-RNNLMs, the previous history $w_1^{t-1}$ is modeled with recurrent units 
(e.g. LSTM, GRU). However, instead of modeling the complete future context information, $w_{t+1}^L$,
using recurrent units, feedforward units are used 
to capture a finite number of succeeding words, $w_{t+1}^{t+k}$.
The softmax function is again applied at the output layer to
obtain the probability of the current word $P(w_t | w_1^{t-1}, w_{t+1}^{t+k})$.
The word embedding in the projection layer are shared for all input words.
When the succeeding words are beyond the sentence boundary,
a vector of 0 is used as the word embedding vector. This is similar to 
the zero padding of the feedforward forward NNLMs at the beginning of each sentence \cite{NNLM}.

As the number of succeeding words is finite and fixed for each word, its succeeding words
can be organized as a $n$-gram future context and used for minibatch mode training as in feedforward NNLMs \cite{NNLM}.
Su-RNNLMs can then be trained efficiently in a similar fashion to uni-RNNLMs in a
spliced sentence bunch mode \cite{chen2016taslp}.

Compared with equations \ref{eqn:uni} and \ref{eqn:bi}, the probability
of word sequence $w_1^L$ can be computed as
\begin{equation}
 P_{s} (w_1^L) = \frac{1}{Z_{s}} \prod_{t=1}^{L} P(w_t | w_1^{t-1}, w_{t+1}^{t+k})
  \label{eqn:succ}
\end{equation}
Again, the sentence level normalization term $Z_{s}$ is difficult to 
compute and only ``pseudo'' PPL can be obtained.
The probabilities of su-RNNLMs are also very sharp, which can be 
seen from the ``pseudo'' PPLs in Table \ref{tab:wersuccrnnppl} in Section \ref{sec:exps}.
Hence, the bi-RNNLM probability smoothing given in Equation \ref{eqn:smooth} is
also required for su-RNNLMs to achieve good performance at evaluation time.

\section{Lattice rescoring}
\label{sec:rnnlmsuccwordslatrescore}
Lattice rescoring with feedforward NNLMs is straightforward \cite{NNLM}
whereas approximations are required for uni-RNNLMs lattice rescoring \cite{liu2016taslp, sundermeyer2015taslp}.
As mentioned in Section \ref{sec:birnnlms}, N-best rescoring has previously been used for bi-RNNLMs.
It is not practical for bi-RNNLMs to be used for lattice rescoring and generation as both the complete previous 
and future context information are required. However, lattices are very useful in many  applications, such as confidence
score estimation \cite{wessel2001confidence}, keyword search \cite{chen2017icassp} 
 and confusion network decoding \cite{CN}.
In contrast, su-RNNLMs require a fixed number of succeeding words,
instead of the complete future context information. 
From Figure \ref{fig:rnnlmsw},
 su-RNNLMs can be viewed as a combination
  of uni-RNNLMs for history information and
 feedforward NNLMs for future context information.
 Hence, lattice rescoring is feasible for su-RNNLMs by
 extending the lattice rescoring algorithm
of uni-RNNLMs by considering additional fixed length future contexts.

\subsection{Lattice rescoring of uni-RNNLMs}
In this paper, the $n$-gram approximation
\cite{liu2016taslp} based approach is used for uni-RNNLMs lattice rescoring.
When considering merging of two paths,
if their previous $n-1$ words are identical, 
the two paths are viewed as ``equivalent'' and can be merged.
This is illustrated in Figure \ref{fig:unilat} for the start node of word $w_4$.
The history information from the best path is kept for the following RNNLM probability computation
and the histories of all other paths are discarded. For example, the path $(w_0, w_2, w_3)$ is kept
and the other path $(w_1, w_2, w_3)$ is discarded given arc $w_4$.

There are two types of approximation involved for uni-RNNLM lattice rescoring, which are the
merge and cache approximations.
The merge approximation controls the merging of two paths.
In \cite{liu2016taslp}, the first path reaching the node was kept and all other 
paths with the same $n$-gram history were discarded irrespective of the associated scores.
This introduces inaccuracies in the RNNLM probability calculation. 
The merge approximation can be improved by keeping the path with the highest accumulated score. 
This is the approach adopted in this work. 
For fast probability lookup in lattice rescoring, $n$-gram probabilities can be cached using $n-1$ words as a key.
A similar approach can be used with RNNLM probabilities. In \cite{liu2016taslp}, RNNLM probabilities were cached based on the previous 
$n-1$ words, which is referred as cache approximation.  Thus a word probability obtained from the cache may be
derived from another history sharing the same $n-1$ previous words.
This introduces another inaccuracy. In order to avoid this inaccuracy yet
maintain the efficiency, the cache approximation used in \cite{liu2016taslp}
is improved by adopting the complete history as key for caching 
RNNLM probabilities. Both modifications yielt small but consistent improvements
over \cite{liu2016taslp} on a range of tasks.

\subsection{Lattice rescoring of su-RNNLMs}
For lattice rescoring with su-RNNLMs, the $n$-gram approximation can be
adopted and extended to support the future word context. In order to handle succeeding words correctly,
paths will be merged only if the following succeeding words are identical.
In this way, the path expansion is carried out in both directions.
Any two paths with the same succeeding words
and $n-1$ previous words are merged. 

\begin{figure}
    \includegraphics[width=8.0cm]{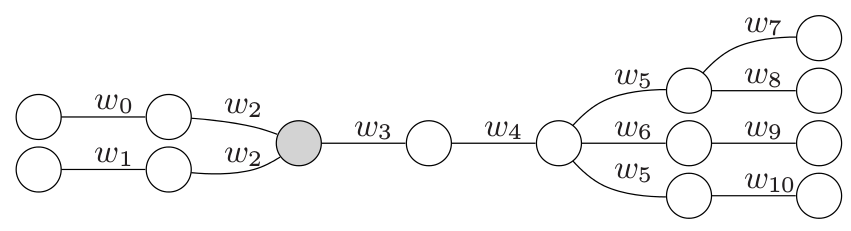}
        \caption{{\it Lattice generated by 2-gram LM.}}
\label{fig:origlat}
  \end{figure}

\begin{figure}
    \includegraphics[width=8.0cm]{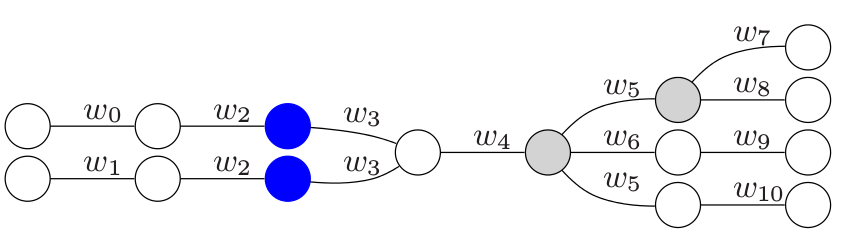}
    \caption{{\it Lattice generated by uni-RNNLMs with 3-gram approximation.}}
\label{fig:unilat}
  \end{figure}

\begin{figure}
    \includegraphics[width=8.0cm]{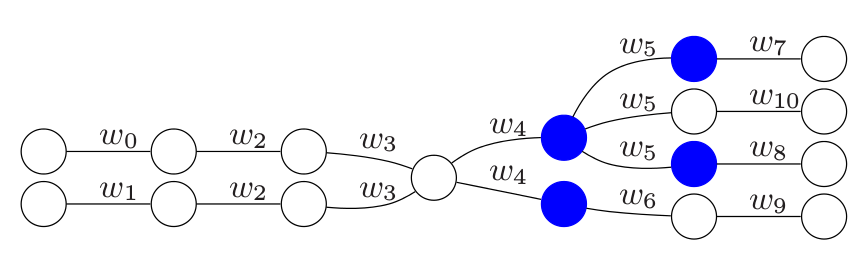}
\caption{{\it Lattice generated by su-RNNLMs with 3-gram approximation for history context and 1 succeeding word.}}
\label{fig:bilat}
  \end{figure}

Figure \ref{fig:origlat} shows part of an example lattice generated by a 2-gram LM.
In order to apply uni-RNNLM lattice rescoring using a 3-gram approximation,
the grey shaded node in Figure \ref{fig:origlat} needs to be duplicated
as word $w_3$ has two distinct 3-gram
histories, which are $(w_0, w_2)$ and $(w_1, w_2)$ respectively.
Figure \ref{fig:unilat} shows the lattice after rescoring using a uni-RNNLM with 3-gram approximation.
In order to apply su-RNNLMs for lattice rescoring, the succeeding words also need to be taken into account.
Figure \ref{fig:bilat} is the expanded lattice using a su-RNNLM with 1 succeeding word. 
The grey shaded nodes in Figure \ref{fig:unilat} need to be expanded further as they have distinct succeeding words.
The blue shaded nodes in Figure \ref{fig:bilat} are the expanded node in the resulting lattice.


Using the $n$-gram history approximation and given $k$ succeeding words, the lattice expansion 
process is effectively a $n+k$-gram lattice expansion for uni-RNNLMs. For larger value
of $n$ and $k$, the resulting lattices can be very large. This can be addressed by pruning the lattice
and doing initial lattice expansion with a uni-RNNLM.

\section{Language Model Interpolation}
\label{sec:intpltlms}
For unidirectional language models, such as $n$-gram model and uni-RNNLMs, 
the word probabilities are normally combined using linear interpolation,
\begin{eqnarray}
\lefteqn{P_u(w_t | w_1^{t-1}) = }\\
&& (1 - \lambda_1) P_{n} (w_t | w_1^{t-1}) +  \lambda_1 P_{r} (w_t | w_1^{t-1})
\nonumber
\end{eqnarray}
where $P_n$ and $P_r$ are the probabilities from $n$-gram and uni-RNN LMs respectively,
$\lambda_1$ is the interpolation weight of uni-RNNLMs.

However, it is not valid to directly combine uni-LMs (e.g unidirectional $n$-gram LMs or RNNLMs)
and bi-LMs (or su-LMs) using linear interpolation 
due to the sentence level normalisation term required for bi-LMs (or su-LMs) in Equation \ref{eqn:bi}.
As described in \cite{xie2017interspeech},
uni-LMs can be log-linearly interpolated with bi-LMs for speech recognition using,
\begin{eqnarray}
\lefteqn{P(w_t | w_1^{t-1}, w_{t+1}^{L}) = }\\
&&  \frac{1}{Z} P_{u} (w_t | w_1^{t-1})^{(1 - \lambda_2)}  P_{b} (w_t | w_1^{t-1}, w_{t+1}^{L})^{\lambda_2}
\nonumber
\end{eqnarray}
where $Z$ is the appropriate normalisation term. The normalisation term can be discarded for speech recognition
as it does not affect the hypothesis ranking. $P_u$ and $P_b$ are the probabilities from 
uni-LMs and bi-RNNLMs respectively. $\lambda_2$ is the log-linear interpolation weight of bi-RNNLMs.
The issue of normalisation term in su-RNLMs is similar to that of bi-RNNLMs,
as shown in Equation \ref{eqn:succ}.
Hence, log-linear interpolation can also be applied for the combination of su-RNNLMs and 
uni-LMs and is the approach used in this paper.

By default, linear interpolation is used to combine uni-RNNLMs and $n$-gram LMs.
A two-stage interpolation is used when including bi-RNNLMs and su-RNNLMs.
The uni-RNNLMs and $n$-gram LMs are first interpolated using linear interpolation.
These linearly interpolated probabilities are then log-linearly interpolated with those of bi-RNNLMs (or su-RNNLMs).

\section{Experiments}
\label{sec:exps}

Experiments were conducted using the AMI IHM meeting corpus~\cite{AMI}
to evaluated the speech recognition performance of various 
language models. The Kaldi training data configuration was used.
A total of 78 hours of speech was used
in acoustic model training. This consists of about 1M words of acoustic transcription.
Eight meetings were excluded from the training set and used 
as the development and test sets. 

The Kaldi acoustic model training recipe \cite{povey2011kaldi} featuring
sequence training~\cite{vesely2013interspeech} was applied for
deep neural network (DNN) training.
CMLLR transformed MFCC features \cite{CMLLR} were used as the input and
4000 clustered context dependent states were used as targets. 
The DNN was trained with 6 hidden layers, and each layer has 2048 hidden nodes.

The first part of the Fisher corpus, 13M words, was used for additional language modeling training data.
A 49k word decoding vocabulary was used for all experiments. 
All LMs were trained on the combined (AMI+Fisher), 14M word in total. A 4-gram KN smoothed back-off LM without 
pruning was trained and used for lattices generation. GRU based recurrent units were used for all unidirectional and bidirectional RNNLMs
\footnote{GRU and LSTM gave similar performance for this task, while GRU LMs are faster for training and evaluation}.
512 hidden nodes were used in the hidden layer. An extended version of CUED-RNNLM \cite{CUEDRNNLM} was developed for the
training of uni-RNNLMs, bi-RNNLMs and su-RNNLMs. The related code and recipe will be available online
\footnote{http://mi.eng.cam.ac.uk/projects/cued-rnnlm/}.
The linear interpolation weight $\lambda_1$ between 4-gram LMs and uni-RNNLMs was set to be 0.75 as it gave the best performance
on the development data.
The log-linear interpolation weight $\lambda_2$ for bi-RNNLMs (or su-RNNLMs) was 0.3.
The probabilities of bi-RNNLMs and su-RNNLMs were smoothed with a smoothing factor 0.7 as suggested in \cite{xie2017interspeech}.
The 3-gram approximation was applied for the history merging of uni-RNNLMs and su-RNNLMs during lattice rescoring and generation \cite{liu2016taslp}.

Table \ref{tab:baseline} shows the word error rates of the baseline system with
4-gram and uni-RNN LMs. Lattice rescoring and $100$-best rescoring are applied 
to lattices generated by the 4-gram LM.
As expected, uni-RNNLMs yield a significant performance
improvement over 4-gram LMs. Lattice rescoring gives a comparable performance 
with 100-best rescoring. Confusion network (CN) decoding can be
applied to lattices generated by uni-RNNLM lattice rescoring
and additional performance improvements can be achieved.
However, it is difficult to apply confusion network decoding to the 100-best 
\footnote{N-best list can be converted to lattice and CN decoding then can be applied, 
but it requires a much larger N-best list, such as 10K used in \cite{liu2016taslp}.}.
\begin{comment}
\begin{table}[htbp]
 \centering
 \begin{tabular}{|l|c||c|c|c|c|c|c|c|c|c|c|}
\hline
 \multirow{2}{*}{LM}& \multirow{2}{*}{rescore}  & \multicolumn{2}{|c|}{dev} & \multicolumn{2}{|c|}{eval}  \\
 \cline{3-6}
       &  	  &Vit 	 & CN 	&  Vit 	& CN	\\
\hline
ng4 	& -	&  24.46	&24.20	&24.68	& 24.44 		\\
\hline
\multirow{2}{*}{\ +uni-rnn}
	&100-best& 22.51 & - & 22.68 & -	 \\
        &lattice &22.49	& 22.33	&  22.53 &22.30		\\
\hline
\end{tabular}
 \caption{Baseline WER results on AMI corpus}
\label{tab:baseline}
\end{table}
\end{comment}

\begin{table}[htbp]
 \centering
 \begin{tabular}{|l|c||c|c|c|c|c|c|c|c|c|c|}
\hline
 \multirow{2}{*}{LM}& \multirow{2}{*}{rescore}  & \multicolumn{2}{|c|}{dev} & \multicolumn{2}{|c|}{eval}  \\
 \cline{3-6}
       &  	  &Vit 	 & CN 	&  Vit 	& CN	\\
\hline
ng4 	& -	&  23.8	&23.5	& 24.2	& 23.9		\\
\hline
\multirow{2}{*}{\ +uni-rnn}
	&100-best& 21.7	& - 	& 22.1 	& -	 \\
        &lattice & 21.7	&  21.5 & 21.9  & 21.7	\\
\hline
\end{tabular}
 \caption{Baseline WER results on AMI corpus}
\label{tab:baseline}
\end{table}

Table \ref{tab:wersuccrnnppl} gives the training speed measured with word per second (w/s) and
(``pseudo'') PPLs of various RNNLMs with difference amounts of future word context. When the number of succeeding words
is 0, this is the baseline uni-RNNLMs. When the number of succeeding words is set to $\infty$,
 a bi-RNNLM with complete future context information is used.
It can be seen that su-RNNLMs give a comparable training speed as 
uni-RNNLMs. The additional computational load of the su-RNNLMs mainly come from
the feedforward unit for succeeding words as shown in Figure \ref{fig:rnnlmsw}.
The computation in this part is much less than that of other parts such 
as output layer and GRU layers. However, the training of su-RNNLMs
is much faster than bi-RNNLMs as it is difficult to parallelise the training of
bi-RNNLMs efficiently \cite{xie2017interspeech}.
It is worth mentioning again that the PPLs of uni-RNNLMs can not 
be compared directly with the ``pseudo'' PPLs of bi-RNNLMs and su-RNNLMs.
But both PPLs and ``pseudo'' PPLs reflect the average log probability of each word.
From Table \ref{tab:wersuccrnnppl}, with increasing number of succeeding words, the 
``pseudo'' PPLs of the su-RNNLMs keeps decreasing, yielding comparable value as bi-RNNLMs.

\begin{table}[htbp]
 \centering
 \begin{tabular}{|c||c|c|c|c|c|c|c|c|c|c|c|}
\hline
\#succ words 	& 0 & 1 &  3   & 7 & $\infty$ 	\\
\hline
train speed(w/s)&4.5K&4.5K&3.9K&3.8K&0.8K\\
(pseudo) PPL 	&66.8&25.5&21.5&21.3&22.4\\
  \hline
\end{tabular}
 \caption{Train speed and (Pseudo) Perplexity of  uni-, bi-, 
 and su-RNNLMs. 0 succeeding word is for uni-RNNLMs and $\infty$ for bi-RNNLMs.}
\label{tab:wersuccrnnppl}
\end{table}

Table \ref{tab:wersuccrnnnbest} gives the WER results of 100-best rescoring with 
various language models. For bi-RNNLMs (or su-RNNLMs), it is not possible to use 
linear interpolation. Thus a two stage approach is adopted as described in Section \ref{sec:intpltlms}.
This results in slight differences, second decimal place, between the uni-RNNLM case
and the 0 future context su-RNNLM.
\begin{comment}
two changes are made compared the use of uni-RNNLMs with linear interpolation. 
An additional bi-RNNLM (or su-RNNLM) is used and a cascade of linear 
and log-linear interpolation is adopted. In order to present a fair comparison and 
validate that the gains are from improved modeling rather than difference in interpolation methods,
the two-step interpolation was also applied for uni-RNNLMs. This involves first linearly interpolating 4-gram LMs and uni-RNNLMs
and then log-linearly interpolated with uni-RNNLMs. 
The WER results are given in the 4th row of Table \ref{tab:wer-succrnn-nbest} with 0 succeeding word.
It can be seen that the su-RNNLMs with 0 succeeding word gives a comparable performance with uni-RNNLMs
using linear interpolation (2nd row).
\end{comment}
The increasing number of the succeeding words consistently reduces the WER. With 1 succeeding word,
the WERs were reduced by 0.2\% absolutely. Su-RNNLMs with more than 2 succeeding words gave about 0.5\% absolute
WER reduction. Bi-RNNLMs (shown in the bottom line of Table \ref{tab:wersuccrnnnbest})
outperform su-RNNLMs by 0.1\% to 0.2\%, as it is able to incorporate the 
complete future context information with recurrent connection.
\begin{comment}
\begin{table}[htbp]
 \centering
 \begin{tabular}{|l|c||c|c|c|c|c|c|c|c|c|c|}
\hline
  \multirow{1}{*}{LM}	& \#succ  words	& dev 	& eval \\
\hline
  ng4 			& -		& 25.00	& 25.19	\\
  \ +uni-rnn		&-		& 22.97	& 23.02	\\
\hline
 \ \ +bi-rnn		&-		&22.49 	& 22.39		\\
\hline
  \multirow{7}{*}{\ \ +su-rnn}
    		 	&0		& 22.95 &23.00 		\\
  		 	&1		& 22.68	&22.80		\\	
			&2		& 22.60	&22.63		\\
			&3		&22.55	&22.63		\\
			&4		&22.56	& 22.57		\\
			&5		&22.54	& 22.53		\\
			&6		&22.52	& 22.53		 \\
			&7		&22.56	& 22.54		\\
  \hline
\end{tabular}
 \caption{WER results of uni-, bi, and su-RNNLMs with 100-best rescoring}
\label{tab:wer-succrnn-nbest}
\end{table}
\end{comment}

\begin{comment}
\begin{table}[htbp]
 \centering
 \begin{tabular}{|l|c||c|c|c|c|c|c|c|c|c|c|}
\hline
  \multirow{1}{*}{LM}	& \#succ  words	& dev 	& eval \\
\hline
  ng4 			& -		& 24.46	& 24.68	\\
  \ +uni-rnn		&-		& 22.51	& 22.68	\\
\hline
 \ \ +bi-rnn		&-		& 21.99	& 22.00	\\
\hline
  \multirow{8}{*}{\ \ +su-rnn}
    		 	&0		& 22.47	& 22.63		\\
  		 	&1		& 22.22	& 22.42		\\	
			&2		& 22.09	&22.27		\\
			&3		&22.05	&22.19		\\
			&4		&22.12	& 22.21		\\
			&5		&22.04	& 22.16		\\
			&6		&22.11	& 22.19		 \\
			&7		&	& 		\\
  \hline
\end{tabular}
 \caption{WER results of uni-, bi, and su-RNNLMs with 100-best rescoring}
\label{tab:wer-succrnn-nbest}
\end{table}
\end{comment}

\begin{table}[htbp]
 \centering
 \begin{tabular}{|l|c||c|c|c|c|c|c|c|c|c|c|}
\hline
  \multirow{1}{*}{LM}	& \#succ  words	& dev 	& eval \\
\hline
  ng4 			& 		& 23.8	& 24.2	\\
  \quad +uni-rnn		&-		& 21.7	& 22.1	\\
% \hline
%  \qquad +bi-rnn		&		& 21.2	& 21.4	\\
\hline
  \multirow{9}{*}{\qquad +su-rnn}
    		 	&0		& 21.7	& 22.1		\\
  		 	&1		& 21.5	& 21.8		\\	
			&2		& 21.3	& 21.7		\\
			&3		& 21.3	& 21.6		\\
			&4		& 21.4	& 21.6		\\
			&5		& 21.3	& 21.6		\\
			&6		& 21.3	& 21.6		 \\
			&7		& 21.4 	& 21.6		\\
			&$\infty$ 	& 21.2	& 21.4		\\
  \hline
\end{tabular}
 \caption{WERs of uni-, bi, and su-RNNLMs with 100-best rescoring. 
 0 succeeding word is for uni-RNNLMs and $\infty$ for bi-RNNLMs.}
\label{tab:wersuccrnnnbest}
\end{table}

Table \ref{tab:wersuccrnnlm} shows the WERs of lattice rescoring using su-RNNLMs.
The lattice rescoring algorithm described in Section \ref{sec:rnnlmsuccwordslatrescore}
was applied. Su-RNNLMs with 1 and 3 succeeding words were used for lattice rescoring.
From Table \ref{tab:wersuccrnnlm}, su-RNNLMs with 1 succeeding words give 0.2\% WER 
reduction and using 3 succeeding words gives about 0.5\% WER reduction. These results are consistent with the
100-best rescoring result in Table \ref{tab:wersuccrnnnbest}. Confusion network decoding can be applied on the 
rescored lattices and additional 0.3-0.4\% WER performance improvements are obtained on dev and eval test sets.
\begin{comment}
\begin{table}[htbp]
 \centering
 \begin{tabular}{|l|c||c|c|c|c|c|c|c|c|c|c|}
\hline
 \multirow{2}{*}{LM} & \#succ & \multicolumn{2}{|c|}{dev} & \multicolumn{2}{|c|}{eval}  \\
 \cline{3-6}
			&words	&Vit 	 & CN 	&  Vit 	& CN	\\
\hline
ng4 		  	&	& 25.00&24.19	&25.19	&  24.44		\\
\hline
\ +uni-rnn 	 	& 	&23.11&22.33	& 22.95	& 22.30	\\
\hline
% 			& 0  	& 23.06 &22.27	&22.96 	& 22.21 \\
\ \ +su-rnn		& 1	&22.92 	&22.12	&22.77	&22.11	\\
			& 3	& 22.67 &21.77	& 22.40	& 21.53	\\
\hline
\end{tabular}
 \caption{WER results of uni-, bi, and su-RNNLMs with lattice rescoring}
\label{tab:wer-succ-rnnlm}
\end{table}
\end{comment}

\begin{comment}
\begin{table}[htbp]
 \centering
 \begin{tabular}{|l|c||c|c|c|c|c|c|c|c|c|c|}
\hline
 \multirow{2}{*}{LM} & \#succ & \multicolumn{2}{|c|}{dev} & \multicolumn{2}{|c|}{eval}  \\
 \cline{3-6}
			&words	&Vit 	 & CN 	&  Vit 	& CN	\\
\hline
ng4 		  	&-	& 24.46	&24.20	&24.68	& 24.44 		\\
\hline
\ +uni-rnn 	 	& -	&22.49	&22.33	& 22.53	& 22.30	\\
\hline
% 			& 0  	&  	&	& 	&  \\
\ \ +su-rnn		& 1	& 22.32	&22.13	&22.20	&22.09	\\
			& 3	& 21.97?&21.77	& 21.70?&21.53	\\
\hline
\end{tabular}
 \caption{WER results of uni-RNNLMs and su-RNNLMs with lattice rescoring}
\label{tab:wer-succ-rnnlm}
\end{table}
\end{comment}

\begin{table}[htbp]
 \centering
 \begin{tabular}{|l|c||c|c|c|c|c|c|c|c|c|c|}
\hline
 \multirow{2}{*}{LM} & \#succ & \multicolumn{2}{|c|}{dev} & \multicolumn{2}{|c|}{eval}  \\
 \cline{3-6}
			&words	&Vit 	 & CN 	&  Vit 	& CN	\\
\hline
ng4 		  	&-	& 23.8	&23.5	&24.2	&  23.9		\\
% \hline
\quad +uni-rnn 	 	& -	& 21.7	&21.5	& 21.9	& 21.7 	\\
\hline
% 			& 0  	&  	&	& 	&  \\
\qquad +su-rnn		& 1	& 21.6	&21.3	& 21.6	& 21.5	\\
			& 3	& 21.3	&21.0	& 21.4	& 21.1	\\
\hline
\end{tabular}
 \caption{WERs of uni-RNNLMs and su-RNNLMs with lattice rescoring}
\label{tab:wersuccrnnlm}
\end{table}

\section{Conclusions}
\label{sec:conclusion}
In this paper, the use of future context information 
on neural network language models has been explored.
A novel model structure is proposed to address the issues associated with bi-RNNLMs,
such as slow train speed and difficulties in lattice rescoring.
Instead of using a recurrent unit to capture the complete future information,
a feedforward unit was used to model a finite number of succeeding words.
The existing training and lattice rescoring algorithms for uni-RNNLMs 
are extended for the proposed su-RNNLMs.
Experimental results show that su-RNNLMs achieved a slightly worse performances than
bi-RNNLMs, but with much faster training speed. Furthermore, additional performance
improvements can be obtained from lattice rescoring and subsequent confusion network decoding.
Future work will examine improved pruning scheme to address the lattice expansion issues 
associated with larger future context.
\begin{comment}
This paper validates that it is beneficial to utilise future context information
for language modeling. Consistent and significant performance
improvements can be achieved over state-of-the-art uni-RNNLMs.
Nevertheless, there are still a lot of work to investigate. 
We found that the pattern of bi-RNNLMs and su-RNNLMs are quite different to standard, uni-RNNLMs
bi-RNNLMs (or su-RNNLMs) have a much sharper probability distribution. 
This explains why we still need all LMs ($n$-gram, uni- and succ-RNNLMs) instead of directly replacing uni-RNNLMs by bi-RNNLMs.
We are also looking for a more sensible way to smooth the probability of su-RNNLMs,
as it is crucial for tasks with high word error rate.
\end{comment}

\vfill\pagebreak
 \bibliographystyle{IEEEbib}
 \bibliography{reference}

\end{document}